\pdfoutput=1
\documentclass[11pt]{article}
\usepackage[preprint]{acl}
\usepackage{times}
\usepackage{latexsym}
\usepackage[T1]{fontenc}
\usepackage[utf8]{inputenc}
\usepackage{microtype}
\usepackage{inconsolata}
\usepackage{graphicx}
\usepackage{amsmath}
\usepackage{array}

\makeatletter
\ifacl@anonymize
  \newcommand{\repourl}{\url{https://anonymous.4open.science/r/nli-hlv-structure-E6F7}}
\else
  \newcommand{\repourl}{\url{https://github.com/oudeis01/nli-hlv-structure}}
\fi
\makeatother

\title{How Much Human Label Variation Does Formal Semantic Structure Explain? \\ Group-Level Effects and Item-Level Ceilings in NLI}

\author{Haram Choi \\
  University of Bremen \\
  \texttt{hachoi@uni-bremen.de}}

\begin{document}
\maketitle

\begin{abstract}
Human label variation in natural language inference (NLI) is
increasingly treated as a signal to be measured rather than as noise to
be removed. We ask whether one candidate source of that signal, formal
semantic structure such as negation, quantification, and monotonicity,
changes how much annotators disagree and what they disagree about. We
tag the SNLI and MNLI items of ChaosNLI, each labeled by 100 annotators,
with a rule-based monotonicity tagger, check the tagger by hand on a
sample of the same items, and answer three questions. At the group
level, hypotheses that are not purely upward monotone attract somewhat
more disagreement, but an error sensitivity analysis shows that this
difference is sensitive to tagger error. At the item level, formal
structure explains only a few percent of the variation and cannot pick
out the items that attract high disagreement. In composition, the kinds
of disagreement recorded by VariErr and LiTEx do not differ detectably
across the formal boundary. Formal structure therefore belongs in the
inventory of disagreement sources, with a small and bounded weight.
ChaosNLI was built from low-agreement items, and every claim holds
within that scope. Analysis decisions were written in a
version-controlled research log before the corresponding results were
computed, and negative results are reported in full.
\end{abstract}

\section{Introduction}
\label{sec:intro}

Natural language inference (NLI) asks whether a hypothesis sentence
follows from a premise. The task has often been built on the assumption
that each item has one correct label, and
\citet{pavlick-kwiatkowski-2019-inherent} tested that assumption
directly: disagreement in inference judgments persists when more
annotators are added and cannot be reduced to noise around a single
true label. ChaosNLI \citep{nie-etal-2020-learn} made this variation
measurable at scale by collecting 100 labels per item, so that each
item's label distribution becomes an object of study. A growing
perspectivist literature now treats such variation as information about
the item and the annotators rather than as error to be removed
\citep{plank-2022-problem,uma-etal-2021-learning}.

If variation is information, the next question is where it comes from.
Some sources lie with the annotators: background knowledge, reading
habits, tolerance for vagueness. Others may lie in the item itself.
This paper examines one item-side source that can be defined without
looking at any label: formal semantic structure. Words such as
\textit{not}, \textit{every}, \textit{few}, \textit{most}, and
\textit{only} change which inferences a sentence supports, and they do
so in ways that formal semantics describes precisely
(Section~\ref{sec:motivation}). If such words make inference harder to
track, items that contain them should attract more disagreement, and
perhaps a different kind of disagreement. If annotators handle them
without trouble, formal structure should leave little trace in the
label distributions. Both outcomes are informative, and to our
knowledge the question has not been measured directly in NLI.

We ask three research questions.

\textbf{RQ1.} Is the amount of disagreement, measured at the group
level, different on the two sides of a formal boundary, namely between
hypotheses whose monotonicity profile is purely upward and all other
hypotheses?

\textbf{RQ2.} How much of the item-level variation in disagreement does
formal structure explain?

\textbf{RQ3.} Is the composition of disagreement, meaning the share of
annotation errors and the types of explanation annotators give,
different on the two sides of the boundary?

We answer them with a rule-based operator and monotonicity tagger
applied to the 3,113 SNLI and MNLI items of ChaosNLI. The tagger is
validated against the MED monotonicity benchmark
\citep{yanaka-etal-2019-neural} and, because MED sentences differ from
ChaosNLI sentences, also checked by hand on a stratified sample of
ChaosNLI hypotheses. For RQ3 we use the 498 items that ChaosNLI shares
with VariErr \citep{weber-genzel-etal-2024-varierr} and its LiTEx
annotations \citep{hong-etal-2025-litex,hong-etal-2026-agree}. The contributions pair with the questions.

\begin{enumerate}
\item For RQ1, a group-level difference: hypotheses outside purely
  upward environments show higher label entropy, and the difference is
  not explained by the mere presence of operators or by sentence length
  in rank-based tests. A hand check and an error sensitivity analysis
  show, however, that the difference is sensitive to tagger error
  (Section~\ref{sec:rq1-sensitivity}).
\item For RQ2, an item-level ceiling: formal features account for only
  a few percent of entropy variance and do not identify
  high-disagreement items.
\item For RQ3, composition invariance: error shares and explanation-type
  shares show no detectable difference across the boundary, in tests
  with high power for effects of the size we consider meaningful.
\end{enumerate}

All claims are conditioned on the sample. ChaosNLI selected SNLI and
MNLI items whose original majority label agreed with only three of five
annotators \citep[Section 3.1]{nie-etal-2020-learn}, so effects here may
differ from those on unrestricted NLI data.

The analysis decisions of this paper, including outcome variables,
tests, multiplicity corrections, a power-based reporting rule, and
interpretation rules, were written in a version-controlled research log
before the corresponding results were computed. The evidence for that
timing is the git history of the author's own repository, which the
author controls, so it is weaker than a record held by a third party.
Section~\ref{sec:decisions} describes what was recorded when, and
Section~\ref{sec:robustness} discloses one interpretation rule that
contained a logical error and was corrected after the results were
seen.

\section{Why Formal Structure Might Shape Disagreement}
\label{sec:motivation}

\begin{table*}[t]
\centering
{\small
\begin{tabular}{>{\raggedright\arraybackslash}p{52pt} >{\raggedright\arraybackslash}p{125pt} >{\raggedright\arraybackslash}p{78pt} >{\raggedright\arraybackslash}p{38pt} r r r r}
\hline
{}Class & Premise & Hypothesis & Trigger & E & N & C & Entropy \\
\hline
{}upward & Tourist Information offices can be very helpful. & Some Tourist Information offices are very unhelpful. & none & 25 & 49 & 26 & 1.510 \\
{}upward & These traditional low-{}drafted craft ply effortlessly and quietly through the water guided by their experienced pilots. & There is a lack of experienced pilots for these low-{}drafted craft. & none & 7 & 45 & 48 & 1.295 \\
{}downward & and uh i know what nothing is when i moved out there & I discovered what nothing was when I moved out there. & nothing & 80 & 13 & 7 & 0.909 \\
{}downward & There was no longer any  when you wanted some unbridled adult fun,{} Las Vegas was the place to be. & Las Vegas was billed as a fun destination for all ages. & all & 6 & 33 & 61 & 1.206 \\
{}non-{}monotone & The commentary is chanted by a chorus of six to eight narrators (reminiscent of the chorus in Greek tragedy) who sit at the side of the stage,{} while musicians positioned at the back of the stage provide stark accompaniment with flute and drums. & The musicians sitting near the stage are the most important part of the show. & most & 3 & 76 & 21 & 0.925 \\
{}non-{}monotone & Joseph Lister pioneered the use of carbolic acid to keep wounds clean,{} and James Young Simpson experimented with chloroform as an anesthetic. & Lister and Simpson were the only ones to use carbolic acid and chloroform for this purpose. & only & 14 & 65 & 21 & 1.274 \\
{}mixed & The results of even the most well designed epidemiological studies are characterized by this type of uncertainty,{} though well-{}designed studies typically report narrower uncertainty bounds around the best estimate than do studies of lesser quality. & Most studies are not planned well. & not,{} most & 6 & 53 & 41 & 1.256 \\
{}mixed & but i think a lot of kids it'{}s funny get the same kind of fears like there'{}s somebody under the bed & I'{}m sure most children don'{}t think it'{}s amusing at all. & n'{}t,{} most & 21 & 57 & 22 & 1.416 \\
\hline
\end{tabular}
}

\caption{Illustrative ChaosNLI items for each hypothesis-side
monotonicity class. Items were drawn from the hand-check sample
(Section~\ref{sec:handcheck}) among those where the human judgment
matched the tagger class and no flag was set, two per class with seed
20260929. They illustrate the
classes and are not representative of them. Trigger is the token that
fired the tagger's trigger, \textit{none} for upward. E, N, and C are
the counts of entailment, neutral, and contradiction labels out of 100.
Entropy is in bits (base 2, maximum 1.585). Text is verbatim from the
source data, including a double space in the fourth premise.}
\label{tab:examples}
\end{table*}

Monotonicity describes how a sentence's truth behaves when one of its
parts is replaced by something more general or more specific. In
\textit{Some dogs bark}, replacing \textit{dogs} with the more general
\textit{animals} keeps the sentence true: \textit{Some animals bark}
follows. This position is \emph{upward} monotone. In \textit{No dogs
bark}, the direction reverses: \textit{No animals bark} does not
follow, but \textit{No puppies bark} does. This position is
\emph{downward} monotone, and negation, \textit{few}, \textit{without},
and the first argument of \textit{every} create such positions
\citep{yanaka-etal-2019-neural}. In \textit{Most dogs bark}, neither
replacement of \textit{dogs} is safe in general; such positions are
\emph{non-monotone} \citep{yanaka-etal-2019-neural}. Formal semantics
treats these directions as a calculus \citep{icard-moss-2014-recent},
and NLI benchmarks such as MED
\citep{yanaka-etal-2019-neural} and HELP
\citep{yanaka-etal-2019-help} test whether models follow it.

Why might this calculus bear on human disagreement? In an upward
environment, the everyday strategy of checking whether the hypothesis
describes the same situation in more general words usually gives the
answer that the calculus gives. In downward and non-monotone
environments, the same strategy can give the opposite answer, and the
correct answer depends on tracking the scope and direction of an
operator. Annotators who track it and annotators who do not may then
split. On this view, marked environments should raise disagreement and
should shift its composition toward logical-structure conflicts and
annotation errors. A competing view is also plausible: operators make
the logical relation more explicit and could reduce the room for
pragmatic reinterpretation, which
\citet{jiang-marneffe-2022-investigating} include among the sources of
NLI disagreement. We state both views as
motivation only; neither is a claim of this paper.

Formal structure has a methodological advantage as a candidate source.
It can be read off the sentence without any label information, by a
tagger whose rules are fixed in advance. An association between such a
property and label variation is therefore not circular, and it can be
bounded from above and below. Table~\ref{tab:examples} shows ChaosNLI
items from each class. The upward examples also show that high
disagreement arises in items without any trigger; the question is how
much the operators add.

\section{Related Work}
\label{sec:related}

\paragraph{Human label variation in NLI.} ChaosNLI
\citep{nie-etal-2020-learn} re-annotated low-agreement SNLI and MNLI
development items with 100 labels each.
\citet{jiang-marneffe-2022-investigating} proposed a taxonomy of
disagreement sources for NLI and framed their identification as a
detection task. LiveNLI \citep{jiang-etal-2023-livenli} added
ecologically collected explanations. VariErr
\citep{weber-genzel-etal-2024-varierr} separated annotation error from
genuine variation on 500 ChaosNLI-MNLI items through a two-round
validation design. LiTEx \citep{hong-etal-2025-litex} introduced a
taxonomy of explanation types for within-label variation and applied it
to e-SNLI; a follow-up extends it to cross-label disagreement and
annotates the VariErr items with it \citep{hong-etal-2026-agree}. Position work argues for
treating NLI ambiguity detection as a first-class task
\citep{jayaweera-dorr-2025-disagreement}. We do not propose a taxonomy
or a detector; we measure how far an independently defined formal
partition of the items aligns with the variation these resources
record.

\paragraph{Monotonicity in NLI.} MED \citep{yanaka-etal-2019-neural}
provides 5,382 items whose gold labels depend on upward, downward, or
non-monotone environments, and HELP \citep{yanaka-etal-2019-help}
provides training pairs generated from the monotonicity calculus. This
line evaluates models against monotonicity; we use its benchmark in the
opposite direction, as an external validation target for a tagger whose
output we then relate to human label variation. The tagger performs
operator-constrained label assignment by adapting the dependency-based
classification pattern of prior work \citep{choi-2026-rhetorical}, which
classified discourse connectives within a bounded dependency radius.
The human validation kappa reported there, between 0.32 and 0.42,
concerns a different construct, a graded rating of analytical tendency,
and does not validate the connective classification or our tagger.

\paragraph{Explaining disagreement from item features.} Recent work in
other tasks predicts or analyzes annotation disagreement from item-level
features. \citet{zhang-coltekin-2026-quantifying} predict item-level
rating variance in offensive, hate, and toxic language perception and
report moderate Spearman correlations with human annotation variance.
\citet{maurer-etal-2026-linguistic} model annotation variation in
offensiveness, hate speech, and toxicity ratings from linguistic item features and annotator
characteristics and conclude that effect patterns vary considerably
across datasets. \citet{abdulmumin-etal-2026-temporal} find that
tweet-level linguistic features show no meaningful association with
inter-annotator agreement in a sentiment corpus, where the temporal
simultaneity of annotation dominates instead. The picture from adjacent
tasks is mixed and dataset-dependent. For NLI, and for formal semantic
features in particular, we are not aware of prior work that measures
their explanatory power for label variation in variance-explained form.

\section{Data and Measurement}
\label{sec:data}

\subsection{Data}
\label{sec:datasets}

We use the SNLI and MNLI-matched portions of ChaosNLI
\citep{nie-etal-2020-learn}: 3,113 items (1,514 SNLI, 1,599 MNLI-m),
each with a distribution over entailment, neutral, and contradiction
from 100 crowd labels. We exclude the AlphaNLI portion, whose abductive
format has no premise-hypothesis pair to analyze for monotonicity. As
noted above, ChaosNLI consists of development items whose
original majority label ``agrees with only three out of five individual
labels'' \citep[Section 3.1]{nie-etal-2020-learn}, so the sample is
restricted toward high disagreement.

For RQ3 we use VariErr \citep{weber-genzel-etal-2024-varierr}, whose 500
MNLI items are all contained in ChaosNLI-MNLI and whose two-round design
marks individual labels as errors or as genuine variation, and the LiTEx
annotations of the VariErr items \citep{hong-etal-2026-agree}, which
classify 1,799 explanation rows into eight explanation categories. The
three-way overlap contains 498 items, all from MNLI-m; two VariErr items
are absent from the LiTEx release.

ChaosNLI's bundled README states a Creative Commons Non-Commercial 4.0
license, MED (Section~\ref{sec:medval}) is released under CC BY-SA 4.0,
LiTEx under Apache 2.0, and the VariErr repository carries no license
file. We use all four datasets for non-commercial research and
redistribute none of them: the reproduction repository contains only
analysis code and a fetch script that retrieves each dataset from its
original source at a pinned version.

\subsection{Operator tagger}
\label{sec:tagger}

The tagger assigns each premise and each hypothesis an operator profile
from a spaCy dependency parse (\texttt{en\_core\_web\_trf}). It covers
four operator families: negation, quantifiers in four subtypes
(universal, existential, numeric, proportional), negative polarity
items, and monotonicity triggers. Monotonicity is approximated
lexically. A closed list of downward triggers (negation, \textit{few},
\textit{little}, \textit{without}, a small adverb list, the restrictor
of universal quantifiers, and \textit{at most}) and non-monotone
triggers (\textit{most}, \textit{many}, \textit{only}, \textit{exactly}
with a numeral, and superlatives and ordinals that modify a noun) each
project a scope from the trigger's dependency subtree.

Each sentence then receives a four-way summary. It is \emph{upward}
when no trigger is present, \emph{downward} or \emph{non-monotone} when
only triggers of that type are present, and \emph{mixed} when both types
co-occur. Every analysis in this paper uses this sentence-level summary.
The tagger is deliberately shallow; it does not handle affixal negation,
the free-choice reading of \textit{any}, idiomatic NPIs, conditional
and interrogative environments, or compositional polarity propagation,
and the Limitations section lists these exclusions. Appendix~\ref{app:features}
lists every feature the tagger produces for the RQ2 models.

\subsection{Validation against MED}
\label{sec:medval}

\begin{table}[t]
\centering
{\small
\setlength{\tabcolsep}{5.25pt}
\begin{tabular}{l r r r}
\hline
{}Tagger & \multicolumn{3}{c}{MED reference} \\
{} & upward & downward & non-monotone \\
\hline
\multicolumn{4}{l}{{}(a) Edit-point readout} \\
{}upward & 1589 & 300 & 13 \\
{}downward & 115 & 2887 & 2 \\
{}non-monotone & 114 & 85 & 277 \\
\hline
\multicolumn{4}{l}{{}(b) Sentence-level readout} \\
{}upward & 1039 & 144 & 6 \\
{}downward & 653 & 3027 & 8 \\
{}non-monotone & 123 & 16 & 276 \\
{}mixed & 3 & 85 & 2 \\
\hline
\end{tabular}
}

\caption{Tagger output against the MED reference class on all 5,382 MED
items. Rows are tagger classes and columns are MED classes. (a) reads
the tagger's direction at the edited position of each MED item. (b) reads
the sentence-level summary that the ChaosNLI analyses use; it has a
mixed row because the summary has a mixed class and MED does not.}
\label{tab:med}
\end{table}

MED \citep{yanaka-etal-2019-neural} assigns each of its 5,382 items an
upward, downward, or non-monotone class.\footnote{The released file
contains 3,272 downward, 1,818 upward, and 292 non-monotone items, while
Table 5 of the MED paper prints 3,270 and 1,820 for the first two. We
use the released file.} A MED class describes the monotonicity of the
edited position, not of the whole sentence. Our primary MED metric
therefore reads the tagger's direction at the edit site: it aligns the
premise and hypothesis, locates the replaced span, and evaluates the
direction at the span's syntactic head. The tagger's rules were revised
twice against MED before the ChaosNLI analyses, and each revision was
written in the research log before its re-measurement. The final tagger
agrees with MED on 0.883 of items at the edit site and on 0.807 with its
sentence-level summary. Table~\ref{tab:med} gives both confusion
matrices. The main cost of the final revision is visible in panel (a):
the non-monotone trigger \textit{only} over-fires into upward items.

\subsection{Hand check on ChaosNLI}
\label{sec:handcheck}

\begin{table*}[t]
\centering
{\small
\begin{tabular}{l r r l r l}
\hline
{}Class & $n$ & \multicolumn{2}{c}{Narrow} & \multicolumn{2}{c}{Broad} \\
{} &  & $k/n$ & $p$ [95\% CI] & $k/n$ & $p$ [95\% CI] \\
\hline
\multicolumn{6}{l}{{}(a) Four-class precision} \\
{}upward & 150 & 149/150 & 0.993 [0.965, 0.999] & 138/150 & 0.920 [0.867, 0.953] \\
{}downward & 80 & 70/80 & 0.875 [0.796, 0.926] & 75/80 & 0.938 [0.872, 0.971] \\
{}non-monotone & 42 & 37/42 & 0.881 [0.778, 0.940] & 36/42 & 0.857 [0.750, 0.923] \\
{}mixed & 28 & 23/28 & 0.821$^\dagger$ & 25/28 & 0.893$^\dagger$ \\
\hline
\multicolumn{6}{l}{{}(b) Binary boundary precision} \\
{}upward & 150 & 149/150 & 0.993 [0.965, 0.999] & 138/150 & 0.920 [0.867, 0.953] \\
{}downward & 80 & 70/80 & 0.875 [0.796, 0.926] & 76/80 & 0.950 [0.888, 0.978] \\
{}non-monotone & 42 & 38/42 & 0.905 [0.807, 0.956] & 39/42 & 0.929 [0.837, 0.970] \\
{}mixed & 28 & 28/28 & 1.000$^\dagger$ & 28/28 & 1.000$^\dagger$ \\
{}Non-upward (pooled) & 150 &  & 0.888 [0.838, 0.938] &  & 0.948 [0.914, 0.982] \\
\hline
\end{tabular}
}

\caption{Tagger precision on the 300-item hand check of ChaosNLI
hypotheses, by tagger class. $k/n$ is the number of items where the
human judgment matches the tagger class. Narrow judgments apply the
tagger's own trigger list; broad judgments apply unrestricted linguistic
judgment (Appendix~\ref{app:broad}). (a) requires an exact match of the
four-way class; (b) requires only a match on the upward versus
non-upward boundary, so its upward row equals the upward row of (a) by
definition. Intervals are 95\% Wilson intervals with a finite population
correction. $^\dagger$The mixed class was sampled in full (28 of 28), so
no interval is given. The pooled non-upward row is a stratified estimate
over the 555 tagged non-upward items, with each stratum weighted by its
population size and a normal-approximation interval with finite
population correction; it therefore differs from the unweighted row sum
(narrow 136/150, broad 143/150), and its $k/n$ cell is left blank. The
five items flagged as uncertain are counted at the annotator's best
judgment.}
\label{tab:handcheck}
\end{table*}

MED validation leaves a gap. MED sentences come from linguistics
examples and crowd rewrites \citep{yanaka-etal-2019-neural}, not from the SNLI and MNLI sources of
ChaosNLI, and MED error rates do not transfer to ChaosNLI. For example, the
sentence-level summary labels 0.428 of MED's upward items as
non-upward; if that rate held on ChaosNLI, at least that share of all
ChaosNLI hypotheses would be tagged non-upward, but the observed share
is 555 of 3,113. We therefore measured the tagger's precision on
ChaosNLI directly.

\paragraph{Design.} We drew a stratified sample of 300 hypotheses by
tagger class (seed 20260924): 150 upward, 80 downward, 42 non-monotone,
and all 28 mixed items. Because the number of items in each tagger class
is known, estimating precision per class is enough to reconstruct an
estimated confusion matrix for the whole sample. Only hypotheses were
checked, since the RQ1 boundary is defined on the hypothesis side; the
premise side is unverified. The single annotator, the author, judged
each hypothesis twice. The \emph{narrow} judgment asks which class the
tagger's own trigger list implies; a narrow disagreement is an
implementation error such as a misparse or a misapplied rule. The
\emph{broad} judgment asks whether the sentence contains a downward or
non-monotone environment on linguistic grounds, regardless of the list;
a difference between broad and narrow judgments is a scope exclusion
that the tagger was never meant to cover. Annotators saw only the
hypothesis text, without tagger output, premise, or labels. This
blinding was a matter of discipline rather than mechanism, because the
item identifiers can be traced back to tagger classes from public
files. Sixty items were annotated again after at least three days, and
the narrow four-way judgments agreed with Cohen's $\kappa = 0.93$.

\paragraph{Result.} Table~\ref{tab:handcheck} gives the precision
estimates. Tagged-upward hypotheses are almost always upward under the
narrow judgment (149 of 150). Tagged non-upward hypotheses are less
clean: the stratified estimate is that about one in nine of them
contains no trigger from the list. Table~\ref{tab:failures} shows one
example for each direction of disagreement, for instance
\textit{little} firing as a downward trigger where it only describes
size. The narrow disagreements are the tagger errors that bear
on RQ1, and Section~\ref{sec:rq1-sensitivity} uses their rates to test
how far the RQ1 result depends on them. Scope exclusions (broad versus
narrow) and a second screening pass with two language models are
reported in Appendices~\ref{app:broad} and~\ref{app:llm}; the
screening pass changed no narrow judgment.

\begin{table}[t]
\centering
{\small
\begin{tabular}{>{\raggedright\arraybackslash}p{53pt} >{\raggedright\arraybackslash}p{43pt} >{\raggedright\arraybackslash}p{48pt} r}
\hline
{}Hypothesis & Tagger (spaCy token) & Human (annotator text) & Rows \\
\hline
{}It will be the biggest ever. & upward (none) & non-{}monotone (biggest) & 1 \\
{}People hiding in bushes to scare little children. & downward (little) & upward (none) & 10 \\
{}The burden is enough for first class mail to handle. & non-{}monotone (first) & upward (none) [uncertain] & 4 \\
{}Ne.weather is a general discussion group,{} not only about weather. & non-{}monotone (only) & mixed (not,{} only) & 1 \\
{}Not many modern cities contain a large array of costumes. & mixed (not,{} many) & downward (not) & 3 \\
{}I had work at Bent Waters,{} but only for a little while. & mixed (little,{} only) & non-{}monotone (only) & 2 \\
\hline
\end{tabular}
}

\caption{Tagger failures in the hand check, one example per direction
of narrow disagreement (tagger class versus human class). Rows gives the
number of hand-check items in that direction (21 in total). The tagger
column shows the class and the spaCy token that fired the trigger; the
human column shows the class and the trigger text the annotator
recorded. [uncertain] marks items the annotator flagged as hard to
parse.}
\label{tab:failures}
\end{table}

\subsection{Outcome variables}
\label{sec:outcomes}

For each item we compute two outcomes from the 100-label distribution.
\emph{Entropy} is the Shannon entropy in bits, which is 0 when all
annotators agree and $\log_2 3 \approx 1.585$ when the three labels are
equally frequent. \emph{Majority margin} is the difference between the
largest and the second-largest label shares. Both are recomputed from
the raw annotation counts and match the released distributions; the 28
items with tied majorities are excluded from majority-based readings.

Two further quantities appear throughout. Cliff's $\delta$
\citep{cliff-1993-dominance} compares two groups: it is the probability
that a random item from the first group has the higher value minus the
probability that it has the lower value, so it ranges from $-1$ to $1$
and is 0 when neither group tends to be higher. In this paper the first
group is always the upward group, so a negative $\delta$ on entropy
means that upward items disagree less. The significance level $\alpha$
is the threshold below which a $p$-value counts as significant; we use
$\alpha = 0.05$ before correction. Appendix~\ref{app:tests} explains why
each test was chosen.

The tagger labels 2,558 hypotheses as upward, 402 as downward, 125 as
non-monotone, and 28 as mixed. The upward class, which contains every
trigger-free sentence, dominates. This imbalance motivates the binary
boundary of Section~\ref{sec:rq1-binary} and the power-based reporting
rule of Section~\ref{sec:rq1-design}.

\subsection{How analysis decisions were recorded}
\label{sec:decisions}

The research log is a dated Markdown file in the project's git
repository. Before each block of analysis was run, its outcome
variables, tests, correction rules, effect-size thresholds, and
interpretation rules were written in that log and committed. Later
changes were logged with their reasons, including one error in an
interpretation rule (Section~\ref{sec:ra-disclosure}). This record has
a clear limit: the log lives in a repository the author controls, and
commit dates can be set by the author, so readers must trust the history
rather than a third-party timestamp. Part of that
history was lost in an accident, disclosed in
Section~\ref{sec:audit}.

The hand check and the sensitivity analysis were added for this
version. Their design, including the sampling rule, the judgment
definitions, the thresholds, and the principle that a failed
sensitivity test weakens the RQ1 claim, was committed in d63e15e
(2026-09-23) before any item was annotated. Later commits changed only
storage locations, output file layout, and the choice of screening
models. One operational
rule, which specifies which variants and bounds decide the sensitivity
test, was written in the log before the sensitivity results were
computed but was committed together with those results in c20a2d7, so
its timing rests on the log alone.

\section{RQ1: Group-Level Amount of Disagreement}
\label{sec:rq1}

\subsection{Design}
\label{sec:rq1-design}

RQ1 asks whether disagreement differs across monotonicity classes. The
design crosses two outcomes (entropy, margin), two sides (hypothesis,
premise), and three strata (pooled, SNLI, MNLI-m), giving twelve
reporting families. Within each family, a Kruskal-Wallis test
\citep{kruskal-wallis-1952-use} first asks
whether the four classes differ at all; its statistic $H$ grows as the
average ranks of the classes move apart. Pairwise Mann-Whitney tests
\citep{mann-whitney-1947-test} then compare the classes two at a time. Four classes give
$4 \times 3 / 2 = 6$ pairs, so each family has six pairwise tests, and
Holm correction \citep{holm-1979-simple} keeps the chance of any false
positive among the six at $\alpha$. Effect sizes are Cliff's $\delta$
with 95\% bootstrap confidence intervals \citep{efron-1979-bootstrap}
(10,000 resamples, fixed seed).

The classes are very unequal in size, and a fixed minimum class size
would be arbitrary. We use a rule based on statistical power instead.
For each pair we compute the power to detect a smallest effect size of
interest of $|\delta| = 0.33$, the boundary between small and medium
effects in the convention attributed to
\citet{romano-etal-2006-appropriate} and documented by
\citet{torchiano-2020-effsize}. Pairs with power below 0.80 are reported
as estimates without a test conclusion. The same convention calls
$|\delta| < 0.147$ negligible, and we use that band below.

\subsection{Four-class result}
\label{sec:rq1-fourclass}

\begin{figure}[t]
\centering
\includegraphics[width=\columnwidth]{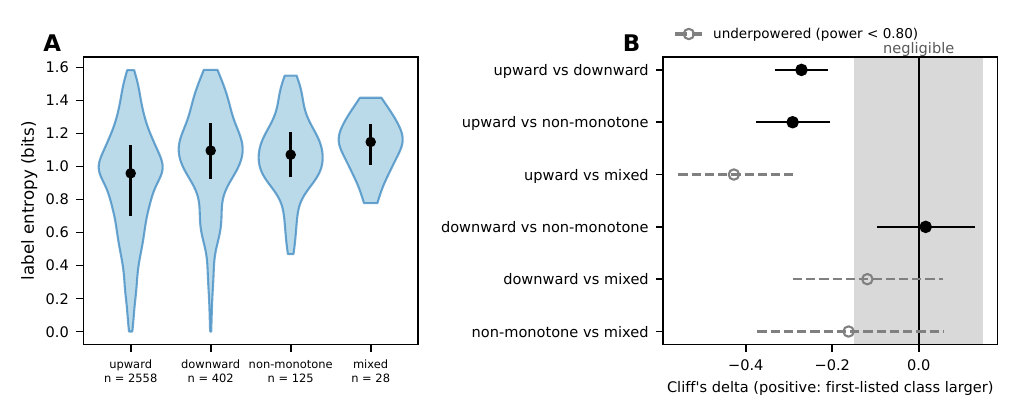}
\caption{Label entropy (bits) across the four hypothesis-side
monotonicity classes, pooled sample ($n = 3{,}113$).}
\label{fig:core-signal}
\end{figure}

Hypothesis-side class is associated with entropy in the pooled sample
(Figure~\ref{fig:core-signal}). Both adequately powered contrasts
against upward are significant and small: $\delta = -0.271$ against
downward and $\delta = -0.292$ against non-monotone. The downward and
non-monotone classes do not differ from each other ($\delta = 0.016$).
The mixed class, with 28 items, falls below the power threshold and is
reported only as an estimate ($\delta = -0.427$). Majority margin shows
the mirror image, with upward items having larger margins than downward
($\delta = 0.201$) and non-monotone ($\delta = 0.195$) items. Premise-side
classes behave like hypothesis-side classes, with all three contrasts
against upward significant.

Two observations follow from the stratified results. First, the
association is concentrated in MNLI-m, where upward versus downward
remains significant but smaller ($\delta = -0.110$); SNLI contains too
few marked hypotheses to carry a test. Because SNLI items are mostly
upward and have lower entropy overall, part of the pooled contrast
reflects the mix of subsets, so every regression below controls for
subset. Second, margin does not reach significance within MNLI-m alone,
so the margin mirror holds only in the pooled sample.

\subsection{The boundary in binary form}
\label{sec:rq1-binary}

Since the three marked classes do not differ from one another, we
reduce the comparison to a binary boundary: purely upward hypotheses
(2,558) versus all others (555). This contrast was decided on after the
four-class result had shown its direction, so we report it as the
effect size for the boundary rather than as a new test. On entropy,
$\delta = -0.284$ (95\% CI $-0.332$ to $-0.235$); on margin,
$\delta = 0.206$. The effect is small in the convention above.

\subsection{Two alternative explanations}
\label{sec:rq1-reductions}

\paragraph{Operator presence.} The upward class is defined by the
absence of triggers, so the boundary might only reflect whether a
sentence contains any operator at all. Within the upward class, we
compare hypotheses with no operator of any kind (2,132) with hypotheses
that contain operators but no trigger (426), for example a numeral or
\textit{some}. If operator presence drove disagreement, the second group
should show higher entropy. It does not: $\delta = 0.007$, with the
whole confidence interval inside the negligible band and power close to
1. A regression of entropy on the twelve operator counts leaves every
count non-significant. The interpretation rule written for this
contrast contained a logical error, disclosed in
Section~\ref{sec:ra-disclosure}.

\paragraph{Length and complexity.} Marked hypotheses might simply be
longer or syntactically deeper. We regress entropy on a marked-hypothesis
indicator and the subset indicator (model M0), then add token counts
and parse-tree depths of both sentences (model M1), with HC3 robust
standard errors \citep{mackinnon-white-1985-some}. The marked coefficient keeps its sign and significance
in both models, at about three hundredths of a bit, against about a
quarter of a bit for the subset indicator. A beta regression that
respects the bounded range of entropy keeps the sign but not the
significance (Section~\ref{sec:robustness}). The defense against the
length explanation therefore rests on the rank-based contrasts more than
on this regression.

\subsection{Sensitivity to tagger error}
\label{sec:rq1-sensitivity}

\begin{figure*}[t]
\centering
\includegraphics[width=\textwidth]{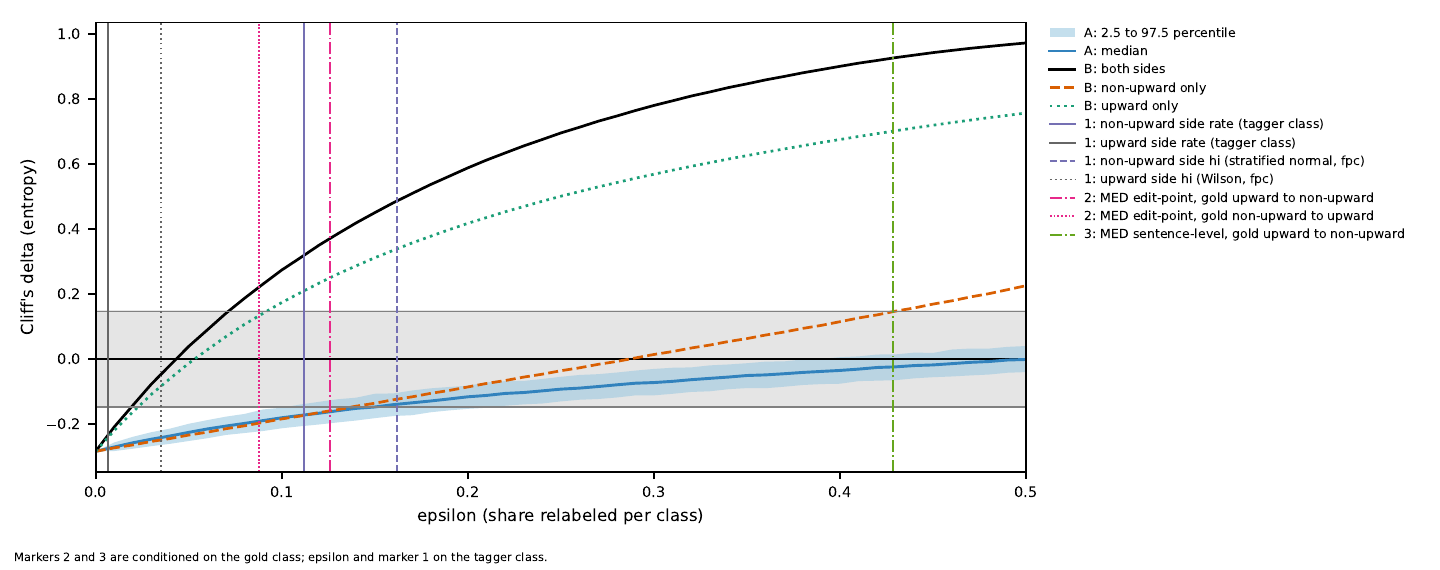}
\caption{Cliff's $\delta$ on entropy for the binary boundary as a share
$\epsilon$ of each class is relabeled (seed 20260924). Analysis A relabels at random (median and 2.5 to 97.5
percentile band over 1,000 repetitions). Analysis B relabels adversarially:
the highest-entropy non-upward items become upward, the lowest-entropy
upward items become non-upward, or both. The grey band is the negligible
range $|\delta| < 0.147$. Vertical lines mark the hand-check error rates
and their 95\% upper bounds (marker 1) and MED error rates (markers 2 and
3). MED markers are conditioned on the gold class, while $\epsilon$ and
marker 1 are conditioned on the tagger class, so they are different
quantities on one axis.}
\label{fig:sensitivity}
\end{figure*}

The hand check measures how often the tagger errs. The question for RQ1
is whether errors at that rate could produce or erase the boundary
effect. Under the narrow judgment, about 11\% of tagged non-upward
hypotheses are upward (95\% upper bound 0.162), and under 1\% of tagged
upward hypotheses are non-upward (upper bound 0.035).

We did not choose a plausible error range and test inside it, since any
chosen range invites the question of why that range. Instead we sweep
the relabeled share $\epsilon$ from 0 to 0.50 and report where the
result breaks down. Analysis A relabels at random. Random errors of
this kind only pull $\delta$ toward zero, so under them the observed
$\delta$ understates the true effect. Analysis B asks the harder
question. It relabels in the way most damaging to the effect, moving
high-entropy non-upward items to upward and low-entropy upward items to
non-upward, which is worse than any error process with the same rate.
Two breakdown points are recorded: where the Holm-corrected $p$ exceeds
0.05, and where $|\delta|$ enters the negligible band. Because the two
measured error rates are very unequal, the decision uses the two
one-sided variants of B, each compared with the upper bound for its
side. The rule fixed in advance was that the RQ1 claim stands if every
breakdown point exceeds its bound and is weakened otherwise
(Section~\ref{sec:decisions}).

Figure~\ref{fig:sensitivity} shows the result. Significance survives:
in both one-sided variants it breaks down only above the measured upper
bounds (0.22 against 0.162, and 0.05 against 0.035). The effect size
does not survive: $|\delta|$ enters the negligible band at 0.14 in the
non-upward variant and at 0.03 in the upward variant, both below the
bounds. Relabeling both sides at once removes significance at
$\epsilon = 0.04$. Following the rule fixed in advance, we weaken the
RQ1 claim. The boundary effect is significant for any one-sided error
process at the measured rates, but an error process at those rates that
is correlated with entropy could shrink it to a negligible size.

\subsection{Answer to RQ1}
\label{sec:rq1-answer}

Hypotheses outside purely upward environments show more disagreement
than upward hypotheses, by a small amount, in the pooled sample. The
difference is not explained by operator presence, and a linear control
for length and parse depth leaves it in place, although a model for the
bounded outcome does not. It is, however, sensitive to tagger error:
the measured error rates are low, but not low enough to rule out that
correlated errors account for most of the effect size. We therefore
read RQ1 as a small group-level association whose size is uncertain,
not as a robust effect.

\section{RQ2: How Much Item-Level Variation Formal Structure Explains}
\label{sec:rq2}

\begin{table*}[t]
\centering
{\small
\begin{tabular}{l r l}
\hline
{}Measure & Value & Model \\
\hline
{}$\epsilon^2$ & 0.036 & Kruskal-Wallis $H/(n-1)$, four classes \\
{}OLS $R^2$ & 0.033 & Entropy on three class dummies (reference upward) \\
{}CV $R^2$ & 0.050 & Linear model, 20 formal features, 5-fold CV \\
{}Spearman $\rho$ & 0.205 & Cross-validated predictions vs. observed entropy \\
{}AUC & 0.606 & Logistic classifier, above-median entropy \\
\hline
\end{tabular}
}

\caption{Item-level explanatory power of formal structure for label
entropy, $n = 3{,}113$. The first two rows use the four hypothesis-side
classes only; the last three use all 20 formal features
(Appendix~\ref{app:features}) without the subset indicator, evaluated on
held-out folds. Above-median entropy defines the binary target for AUC.}
\label{tab:rq2}
\end{table*}

RQ1 concerns group medians. A group difference can coexist with a large
overlap between groups, so RQ2 asks how much of the item-to-item
variation in entropy formal structure accounts for.
Table~\ref{tab:rq2} gives five measures. The four-class partition
explains between 3.3 and 3.6\% of entropy variance, whether measured
through the Kruskal-Wallis statistic ($\epsilon^2 = H/(n-1)$;
\citealp{tomczak-tomczak-2014-need}) or through
an ordinary least squares fit on class indicators. Adding every operator
count on both sides does little better under five-fold
cross-validation. A classifier for above-median entropy reaches an AUC
of 0.606, where 0.5 is chance and 1 is perfect separation. These values
were reported as values from the start, without a pass or fail
threshold, because no threshold could be derived from prior work.

For comparison, \citet{zhang-coltekin-2026-quantifying} report Spearman
correlations of roughly 0.3 to 0.45 when predicting rating variance from
richer feature sets in offensive-language tasks; our cross-validated
correlation of 0.205 comes from deliberately narrow features on a
range-restricted sample.

\paragraph{Answer to RQ2.} Formal structure explains a small share of
item-level variation, a few percent, and does not identify which items
will attract high disagreement. The features sit above chance and far
below practical use.

\section{RQ3: Composition of Disagreement}
\label{sec:rq3}

\begin{table*}[t]
\centering
{\small
\begin{tabular}{l r r l r r}
\hline
{}Contrast & $n$ (up/marked) & $\delta$ & 95\% CI & Holm $p$ & Power \\
\hline
{}C1 & 226/107 & $-$0.076 & [$-$0.189, 0.039] & 0.55 & 0.993 \\
{}C2 & 348/150 & $-$0.029 & [$-$0.113, 0.049] & 0.95 & 1.000 \\
{}C3 & 348/150 & 0.027 & [$-$0.078, 0.131] & 0.95 & 1.000 \\
\hline
\end{tabular}
}

\caption{Composition contrasts on the ChaosNLI, VariErr, and LiTEx
overlap, upward versus marked hypotheses. C1: share of an item's
first-round VariErr labels judged errors in the second round, among
items with more than one first-round label. C2: share of an item's
LiTEx explanations in the Logic Conflict category. C3: share in
Pragmatic, Inferential Knowledge, or Factual Knowledge. Category names
follow \citet{hong-etal-2025-litex}; the released annotations call the
first three Logical Structure Conflict, Pragmatic-Level Inference, and
World-Informed Logical Reasoning. $\delta$ is Cliff's $\delta$ with upward as the first
group; intervals are bootstrap intervals; $p$ is Holm-corrected over the
three contrasts. Power is for $|\delta| = 0.33$ at the Holm bound;
values above 0.9995 display as 1.000.}
\label{tab:rq3}
\end{table*}

\begin{figure}[t]
\centering
\includegraphics[width=\columnwidth]{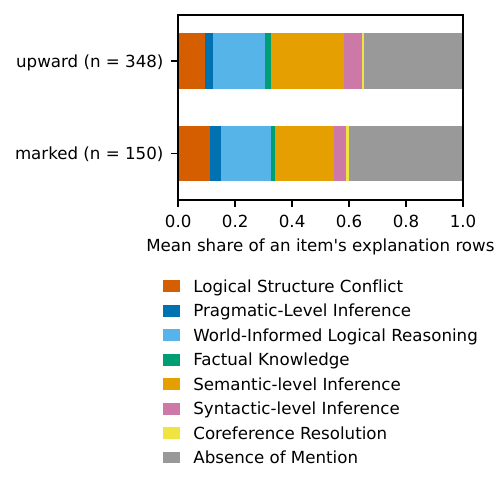}
\caption{Mean share of each LiTEx explanation category per item, for
upward and marked hypotheses in the 498-item overlap.}
\label{fig:composition}
\end{figure}

RQ1 and RQ2 concern the amount of disagreement. RQ3 asks whether its
kind differs across the boundary. If marked environments make
annotators split over logical scope, disagreement on marked items
should contain more annotation errors and more explanations that cite
logical structure, while disagreement on upward items should lean toward
pragmatic and world-knowledge explanations. We tested these three
expectations on the 498 items shared with VariErr and LiTEx, split into
348 upward and 150 marked hypotheses. Each contrast had a directional
expectation written in advance but was tested two-sided, with Holm
correction over the three.

All three contrasts are null (Table~\ref{tab:rq3}), and each has power
of at least 0.99 for an effect at the smallest size of interest. The
intervals for C2 and C3 lie inside the negligible band, and the interval
for C1 stays well short of the smallest effect of interest.
Figure~\ref{fig:composition} shows the same picture across all eight
LiTEx categories: the two groups have similar mixtures.

Three caveats apply. The C2 outcome is mostly zero in both groups, so a
rank test is insensitive to it; Section~\ref{sec:robustness} adds a
test on the share of items with any nonzero value. The three contrasts
share one sample. And the overlap is a non-random subset of the main
sample, all MNLI-m and all selected by VariErr.

\paragraph{Answer to RQ3.} As far as these measures capture it, the
composition of disagreement does not differ across the formal boundary.
Marked and upward items show the same mixture of error, logical-structure
conflict, and pragmatic inference.

\section{Robustness}
\label{sec:robustness}

\begin{table*}[t]
\centering
{\small
\begin{tabular}{l r r r r}
\hline
\multicolumn{5}{l}{{}(a) Zero inflation (C-R1): items with nonzero share} \\
{}Outcome & Upward $k/n$ & Marked $k/n$ & Odds ratio & Holm $p$ \\
\hline
{}VariErr error share & 74/226 (32.7\%) & 43/107 (40.2\%) & 0.725 & 0.66 \\
{}LiTEx logical-structure-conflict share & 72/348 (20.7\%) & 35/150 (23.3\%) & 0.857 & 0.87 \\
{}LiTEx pragmatic and world-knowledge share & 183/348 (52.6\%) & 73/150 (48.7\%) & 1.170 & 0.87 \\
\hline
\multicolumn{5}{l}{{}(b) Bounded-outcome model (C-R2): coefficient on marked hypothesis} \\
{}Model & OLS coef (HC3 SE) & OLS $p$ & Beta coef (SE) & Beta $p$ \\
\hline
{}M0 & 0.036 (0.014) & 0.009 & 0.080 (0.043) & 0.064 \\
{}M1 & 0.031 (0.014) & 0.025 & 0.061 (0.044) & 0.167 \\
\hline
\end{tabular}
}

\caption{Two robustness checks. (a) Items with a nonzero share of each
composition outcome, upward versus marked; the odds ratio is the odds of
a nonzero share in the upward group over the odds in the marked group,
and $p$ is from two-sided Fisher exact tests with Holm correction over
three. (b) Coefficient on the marked-hypothesis indicator in the length
and complexity models of Section~\ref{sec:rq1-reductions}. OLS
coefficients are in bits of entropy; beta regression coefficients are on
the logit scale of rescaled entropy, so the two columns are not
comparable in size.}
\label{tab:robustness}
\end{table*}

\subsection{Five checks}

Five robustness checks were specified in the research log before any of
their statistics were computed. Two facts about that specification are
disclosed first. The numerators of the first check's error-share
comparison were already visible in earlier logged counts, so that
comparison is a sensitivity check rather than a new prediction. And the
input list contained a defect, since two of the counts could not be
computed from the listed files; a correction was logged before execution
with no statistic computed in between.

\paragraph{C-R1, zero inflation.} Table~\ref{tab:robustness}(a) compares
the share of items with any nonzero composition outcome. All three
comparisons are null, so the RQ3 result is not an artifact of zero
inflation.

\paragraph{C-R2, bounded outcome.} Entropy is bounded, so we refit the
length and complexity models as beta regressions after rescaling entropy
to the unit interval and compressing it away from the endpoints
\citep{smithson-verkuilen-2006-better}. The rule was that the linear
models would be considered adequate only if the marked coefficient kept
its sign at $p < 0.05$ in both beta models. The sign is kept but
significance is not (Table~\ref{tab:robustness}(b)). The criterion fails,
and the regression form of the length defense is therefore sensitive to
the distributional model. One asymmetry, not anticipated in the
specification, qualifies the comparison: the linear models use robust
standard errors and the beta models use model-based ones.

\paragraph{C-R3, false discovery rate.} Applying the
Benjamini-Hochberg procedure \citep{benjamini-hochberg-1995-controlling}
instead of Holm changes no conclusion, either for the eleven primary
contrasts or for all 79 contrasts in the analysis plan. Some contrasts
that lack a test conclusion under the power rule, most visibly upward
versus mixed, would count as discoveries if that rule were waived.

\paragraph{C-R4, untagged environments.} The tagger does not process
conditional and comparative environments. Items containing \textit{if}
or \textit{than} make up 7.4\% of the sample, are more common among
marked hypotheses, and have higher median entropy than other items. The
RQ1 boundary is measured with this blind spot included, and we do not
know in which direction full coverage would move it.

\paragraph{C-R5, the mixed class.} The 28 mixed hypotheses mostly
combine sentential negation with a proportional or focus word, as in
\textit{not many} or \textit{not only}, and no combination occurs more
than three times. The class is a small heterogeneous mixture rather than
a natural kind, which is another reason not to read its large point
estimate.

\subsection{A corrected interpretation rule}
\label{sec:ra-disclosure}

The interpretation rule written for the operator-presence contrast
(Section~\ref{sec:rq1-reductions}) was inverted. It stated that a
significant difference between operator-bearing and operator-free
upward items would weaken the operator-presence explanation, and that a
null result would force us to accept that explanation. The logic runs
the other way. If operator presence drove disagreement, upward items
with operators should show higher entropy; a null result therefore
counts against the operator-presence explanation, and a positive
difference would have supported it. The error was noticed and corrected
after the null result had been seen, and the correction was logged with
its timing. The rule's truth table can be checked without data, and the
item-level limits of our claims were already fixed by other results.
Readers who reject any rule correction made after seeing results should
treat this contrast as exploratory; the operator-count regression of
Section~\ref{sec:rq1-reductions}, specified separately, points the same
way.

\subsection{Audit trail}
\label{sec:audit}

Every number in this paper maps to a generating script and a derived
data file in the reproduction repository (\repourl). One event affects
the audit trail. On 2026-07-07 the project directory was accidentally
deleted and restored the same day, and every logged effect size was
reproduced exactly. The git commit objects created after the second
commit of analysis decisions were lost, however, so for decisions
recorded after that point the timing evidence rests on dated session
transcripts rather than on commit hashes. The repository's audit
document records the timeline, the accident, the recovery, and its
verification.

\section{Discussion}
\label{sec:discussion}

\paragraph{For perspectivist NLP.} NLI has often been treated as a task
with one correct answer per item
\citep{pavlick-kwiatkowski-2019-inherent}. Our results are consistent with the view that NLI disagreement is real
and structured, but they place most of that structure outside formal
semantics. On this sample, whether a hypothesis contains a monotonicity
trigger is associated with at most a small shift in how much annotators
disagree, and with no detectable shift in the errors and explanation
types behind the disagreement. For work that models or collects soft
labels in NLI, this suggests that formal operators are not where most
of the variation lives, and that other item properties, such as the
world knowledge and pragmatic reading that LiTEx explanations cite at
similar rates on both sides of the boundary
(Figure~\ref{fig:composition}), deserve the larger share of
attention. This is an inference from one
range-restricted sample.

\paragraph{For formal semantic hypotheses.} Section~\ref{sec:motivation}
set out two views: marked environments might raise disagreement by
making inference harder to track, or lower it by making the logical
relation explicit. The data give weak support to the first view on
amount and none on composition. If annotators split over scope and
direction in marked items, we would expect more logical-structure
explanations there, and we do not find them. Two readings remain
open, and our data cannot separate them. Crowd annotators may not
engage the monotonicity calculus at all when they judge ordinary
sentences, so that marked items are hard for the same pragmatic reasons
as other items (speculation). Alternatively, a lexical tagger may miss
the compositional structure that actually matters, so that a better
instrument would find more. A compositional tagger, or items built to
vary only in monotonicity, would distinguish the two.

\paragraph{For modeling.} The item-level ceiling has a practical
meaning. Formal features raise the prediction of high disagreement only
modestly above chance, so they cannot replace collecting several labels
per item, and a model that predicts disagreement from them alone would
mostly fail on individual items. They may still serve as a small, cheap,
and interpretable component alongside richer features, whose value
would have to be measured.

\section{Conclusion}
\label{sec:conclusion}

We asked how much of NLI label variation formal semantic structure
explains, and answered in terms of bounds. Hypotheses outside purely
upward environments show slightly more disagreement as a group, but the
size of that difference is sensitive to tagger error. Item by item,
formal structure explains only a few percent of the variation. And the
composition of disagreement, measured through validated error shares and
explanation types, does not differ across the boundary. Formal semantic
structure, in this sample, may shift how much annotators disagree by a
small amount and does not detectably change what they disagree about.

We read this as a calibration result for the perspectivist program:
formal structure belongs in the inventory of disagreement sources with
a small, bounded weight, and most of human label variation in NLI
remains to be explained by other item properties and by annotator-side
variables. Two open questions follow, and we flag both as speculation:
how the boundary behaves on unrestricted samples, where the
low-agreement selection no longer applies, and whether compositional
rather than lexical monotonicity tagging raises the item-level ceiling.
The research log, scripts, and hand-check data released with this paper
are meant to make such extensions easy to run and to check.

\section*{Limitations}

\begin{enumerate}
\item Range restriction. ChaosNLI-S/M contains only development items
  whose original majority label ``agrees with only three out of five
  individual labels'' \citep[Section 3.1]{nie-etal-2020-learn}. All
  effects are within-sample associations over low-agreement items and
  may be attenuated or otherwise distorted relative to unrestricted NLI
  data.
\item Task and language. Everything here is NLI over English sentences;
  nothing licenses transfer to other tasks or languages.
\item Structure versus annotator behavior. We observe no annotator-level
  behavior. An association between sentence structure and label
  dispersion cannot distinguish properties of the items from properties
  of the annotator population that encountered them.
\item Tagger validity. On MED the tagger agrees with the reference on
  0.883 of items at the edit site and 0.807 at the sentence level that
  our analyses use. On ChaosNLI hypotheses, narrow precision is 0.993
  for tagged-upward items and 0.888 for tagged non-upward items
  (stratified estimate). These values come from a single annotator, the
  author, whose blinding was self-imposed; intra-annotator agreement was
  high, but no second annotator was involved. The premise side was not
  checked. The hand check and the sensitivity analysis test the RQ1
  boundary only; the RQ2 and RQ3 results use the same tagger and were
  not tested for sensitivity to its errors.
\item Bounded outcome. Entropy is bounded above by $\log_2 3$, so the
  linear regression coefficients are read for direction and relative
  size only. The beta regression check keeps the sign but not the
  significance of the boundary coefficient, so the regression-based
  defense inherits this limitation; the rank-based results do not.
\item Non-independence in RQ3. The three contrasts share one 498-item
  sample, which is itself a non-random subset (all MNLI-m, all selected
  by VariErr) of the main sample.
\item Residual length signal. Premise token count is borderline
  significant in the M1 regression ($p = 0.047$); length effects are
  controlled, not absent.
\item Genre is uncontrolled. The subset indicator (MNLI-m versus SNLI)
  is the largest coefficient in every regression we fit, and genre was
  not stratified within MNLI; the boundary effect is estimated within,
  not across, this large genre gap.
\item No causal claims. Every result is an observed association in a
  fixed corpus. We make no claim that formal structure causes
  disagreement.
\end{enumerate}

\bibliography{refs}

\appendix

\section{Choice and Interpretation of Tests}
\label{app:tests}

\paragraph{Rank-based tests for entropy and margin.} Both outcomes are
bounded, skewed, and heavily tied, since they are computed from counts
out of 100, and the class sizes are very unequal. Rank-based tests
compare groups through the ordering of values and make no normality
assumption. The Kruskal-Wallis test \citep{kruskal-wallis-1952-use} asks whether any of the four classes
tends to have higher values than the others; its statistic $H$ is a
scaled sum of squared deviations of the class mean ranks from the
overall mean rank. The Mann-Whitney test \citep{mann-whitney-1947-test} does the same for two groups
and is used for the pairwise follow-ups. A small $p$ says that one group
tends to have higher values; it does not say by how much, which is the
job of the effect size.

\paragraph{Cliff's $\delta$.} Cliff's $\delta$
\citep{cliff-1993-dominance} is the effect size that matches a rank
test. It is the difference between the probability that a random item
from the first group exceeds a random item from the second and the
probability of the reverse. It needs no distributional assumption and
reads directly as a dominance probability: $\delta = -0.28$ means that
an upward item has the lower entropy in a random pair noticeably more
often than the higher. We use the thresholds 0.147, 0.33, and 0.474 for
negligible, small, and medium effects attributed to
\citet{romano-etal-2006-appropriate}; the primary source was not
accessible to us, and we take the values from the effsize package
documentation \citep{torchiano-2020-effsize}, which cites it.

\paragraph{Holm and Benjamini-Hochberg corrections.} Running six
pairwise tests raises the chance that at least one is significant by
accident. Holm's procedure \citep{holm-1979-simple} controls that
chance, the family-wise error rate, at $\alpha$ and is uniformly more
powerful than the Bonferroni correction: it compares the smallest
$p$-value with $\alpha/m$, where $m$ is the number of tests, the next
with $\alpha/(m-1)$, and so on. The Benjamini-Hochberg procedure
\citep{benjamini-hochberg-1995-controlling} controls a weaker quantity,
the expected share of false discoveries among the rejections; we use it
only as a sensitivity check (C-R3), to see whether our conclusions
depend on the stricter correction.

\paragraph{Power-based reporting rule.} A non-significant test on a
small group says little, because the test may have had no chance of
detecting a meaningful effect. For each contrast we compute the power
to detect $|\delta| = 0.33$ at the Holm bound given the observed group
sizes, and we give a test conclusion only where that power is at least
0.80. Below it we report the estimate and its interval. Power comes from
the normal approximation for the Mann-Whitney test of
\citet{noether-1987-sample},
$1-\beta = \Phi\bigl(|p - \tfrac{1}{2}|\sqrt{12 n_1 n_2/(n_1+n_2)} - z_{1-\alpha/2}\bigr)$,
where $n_1$ and $n_2$ are the group sizes and $p = (\delta+1)/2$ is the
probability that an item from the first group exceeds one from the
second, with ties counted as one half. The approximation replaces the
variance under the alternative with the null variance and was derived
for continuous data, so it makes no correction for ties and can
overstate or understate power \citep{happ-etal-2019-optimal}.

\paragraph{Bootstrap intervals.} Confidence intervals for $\delta$ are
percentile intervals over 10,000 bootstrap resamples
\citep{efron-1979-bootstrap} of each group, with a fixed seed. They show the range of effect sizes compatible with the data and
are read against the negligible band.

\paragraph{Fisher exact test.} The zero-inflation check (C-R1) compares
two proportions in a 2 by 2 table with some small cells, where the exact
test \citep{fisher-1935-logic} is preferable to a large-sample approximation. The odds ratio is
its effect size.

\paragraph{Regression models.} The length and complexity models (M0, M1)
are linear regressions with HC3 heteroskedasticity-consistent standard
errors \citep{mackinnon-white-1985-some}, because the residual variance of a bounded outcome is unlikely
to be constant. The beta regression check (C-R2) models the rescaled
outcome directly on the unit interval with a logit link
\citep{ferrari-cribari-neto-2004-beta}, after the
compression of \citet{smithson-verkuilen-2006-better} that moves exact
0 and 1 values inside the interval.

\paragraph{Hand-check intervals and agreement.} Per-class precision
intervals are Wilson score intervals \citep{wilson-1927-probable}.
Because the sample is a sizable fraction of each class, we apply a
finite population correction by replacing the sample size $n$ with the
effective sample size $n(N-1)/(N-n)$, where $N$ is the class size,
following the effective sample size approach for proportions from
complex samples \citep{franco-etal-2019-comparative}. A class checked in
full has no sampling error and gets a point estimate only. The pooled
non-upward precision weights each stratum $h$ by its population share
$W_h = N_h/\sum_k N_k$. Its variance is the stratified sampling variance
$\sum_h W_h^2 \frac{N_h-n_h}{N_h-1} \frac{p_h(1-p_h)}{n_h}$
\citep[p.~92]{cochran-1977-sampling}, with the sample proportions $p_h$
in place of the population ones, and its interval uses a normal
approximation.
Intra-annotator agreement is Cohen's $\kappa$ \citep{cohen-1960-coefficient}, which discounts the
agreement expected by chance from the marginal class frequencies.

\section{The 20 Formal Features}
\label{app:features}

The RQ2 cross-validated models (Table~\ref{tab:rq2}) use 20 features,
ten for the premise and ten for the hypothesis. The subset indicator is
excluded. All features come from the tagger of Section~\ref{sec:tagger}.

\begin{table}[h]
\centering
\small
\begin{tabular}{>{\raggedright\arraybackslash}p{56pt} >{\raggedright\arraybackslash}p{139pt}}
\hline
Feature & Definition (counted per sentence) \\
\hline
Negation count & Tokens with the dependency label \texttt{neg} (e.g.\ \textit{not}, \textit{n't}, \textit{never}), plus \textit{no}, \textit{none}, \textit{nobody}, \textit{nothing}, \textit{nowhere}, \textit{neither}, \textit{nor}, \textit{no one} \\
Universal quantifiers & \textit{all}, \textit{every}, \textit{each}, \textit{both} \\
Existential quantifiers & \textit{some}, \textit{somebody}, \textit{someone}, \textit{something}, \textit{several}, \textit{a few} \\
Numeric quantifiers & Tokens with the dependency label \texttt{nummod} \\
Proportional quantifiers & \textit{most}, \textit{many}, \textit{few}, \textit{little}, \textit{much} \\
NPI count & \textit{any}, \textit{anybody}, \textit{anyone}, \textit{anything}, \textit{anywhere}, \textit{ever}, \textit{yet}, \textit{at all}, and sentence-final \textit{either}, licensed or not \\
Class indicators & Four indicators, one per summary class (upward, downward, non-monotone, mixed); exactly one is 1 \\
\hline
\end{tabular}
\caption{Features per sentence. Six counts and four class indicators
for each of the premise and the hypothesis give 20 features. The class
indicators are kept in full rather than dropping one, since the models
are used for prediction rather than coefficient estimation.}
\label{tab:features}
\end{table}

\section{Broad Judgments and the Re-Judgment Pass}
\label{app:broad}

\paragraph{Broad judgments.} The broad judgment applies linguistic
judgment without the trigger list, so conditional antecedents,
questions, comparative clauses, affixal negation, and implicitly
negative predicates such as \textit{doubt} or \textit{refuse} count as
marked. Table~\ref{tab:handcheck} gives broad precision. The broad and
narrow judgments differ on an estimated 7.5\% of the 3,113 hypotheses
(stratified estimate). Most differences run from a narrow upward
judgment to a broad downward one, that is, environments the tagger was
never meant to cover. These are scope exclusions already listed in the
Limitations, not implementation errors, and we do not use them in the
sensitivity analysis. Broad four-way judgments agreed with
$\kappa = 0.77$ on the 60 re-annotated items.

\paragraph{Re-judgment.} After the first-pass judgments were frozen, 71
items selected by the screening pass of Appendix~\ref{app:llm} were
judged again, with the screening outputs visible but without the first
judgment or the stratum. None of the 71 narrow judgments changed. The
first-pass judgments remain the basis of every estimate in the paper.

\section{The Language-Model Screening Pass}
\label{app:llm}

As a check on a single annotator, two language models judged the same
300 hypotheses under the narrow definition, with the same trigger list
and without the human judgments. Their role was restricted in advance to
screening: they select items for a second human look, and they are not
evidence of reliability. The models were Gemma 4 12B
(\texttt{google/gemma-4-12B-it}, revision \texttt{707f0a3}, FP8 dynamic
quantization, vLLM 0.30.0) and gpt-oss-20b
(\texttt{ggml-org/gpt-oss-20b-GGUF}, revision \texttt{ef9b12f}, native
MXFP4, llama.cpp server build b8766), chosen from different model
families so that their errors would be less correlated. Both ran on a
single RTX 4080 at temperature 0 with a fixed seed. The specified runs
used Gemma with thinking disabled and gpt-oss at its default reasoning
effort. Items where either model differed from the human judgment or
gave no answer were selected (48 items). Two further runs with changed
reasoning settings (Gemma with thinking enabled, gpt-oss at low effort)
added 23 items after the fact, giving the 71 re-judged items of
Appendix~\ref{app:broad}. Raw model outputs and run metadata are in the
repository.

\end{document}